\documentclass[sigconf]{acmart}

\RequirePackage[]{url}

\usepackage{hyperref} 
\usepackage{graphicx} 
\usepackage{verbatim} 
\usepackage{multirow} 
\usepackage{subfigure} 

\usepackage[T1]{fontenc} 

\usepackage[utf8]{inputenc} 

\usepackage{microtype} 
\usepackage{tipa} 

\newcommand{\uwa}{The University of Western Australia}
\newcommand{\google}{Google}
\newcommand{\ben}{Hutchinson}
\newcommand{\celeste}{Rodríguez Louro}
\newcommand{\glenys}{Collard}

\newcommand{\eg}{\emph{e.g.}}

\newcommand{\etc}{\emph{etc}}

\newcommand{\redacted}[1]{REDACTED}
\newcommand{\redactedparens}[1]{(REDACTED)}

\renewenvironment{quote}{%
  \list{}{%
    \leftmargin0.69cm   
    \rightmargin\leftmargin
  }
  \item\relax
}
{\endlist}

\title[Designing Speech Technologies for Australian Aboriginal English
]{%
Designing Speech Technologies for Australian Aboriginal English: 
Opportunities, Risks and Participation
}

\author{Ben Hutchinson}
\affiliation{%
  \institution{Google Research}
  \country{Australia} 
}

\author{Celeste Rodríguez Louro}
\affiliation{%
  \institution{University of  
  Western Australia}
  \country{Australia} 
}

\author{Glenys Collard}
\affiliation{%
  \institution{University of  
  Western Australia}
  \country{Australia} 
}

\author{Ned Cooper}
\affiliation{%
  \institution{Australian National 
  University}
  \country{Australia} 
  }

\makeatletter
\let\@authorsaddresses\@empty
\makeatother

\copyrightyear{2025}  
\acmYear{2025}  
\setcopyright{cc} 
\setcctype{by} 
\acmConference[FAccT '25]{The 2025 ACM Conference on Fairness, Accountability, and Transparency}{June 23--26, 2025}{Athens, Greece} \acmBooktitle{The 2025 ACM Conference on Fairness, Accountability, and
Transparency (FAccT '25), June 23--26, 2025, Athens, 
Greece}\acmDOI{10.1145/3715275.3732010} 
\acmISBN{979-8-4007-1482-5/2025/06}

\begin{document}

\begin{abstract} 
In Australia, post-contact language varieties, including creoles and local varieties of international languages, emerged as a result of forced contact between Indigenous communities and English speakers. These contact varieties are widely used, yet are poorly supported by language technologies. This gap presents barriers to participation in civil and economic society for Indigenous communities using these  varieties, and reproduces minoritisation of contemporary Indigenous sociolinguistic identities. This paper concerns three questions regarding this context. First, can speech technologies support speakers of Australian Aboriginal English, a local indigenised variety of English? Second, what risks are inherent in such a project? Third, what technology development practices are appropriate for this context, and how can researchers integrate meaningful community participation in order to mitigate risks? We argue that opportunities do exist---as well as risks---and demonstrate this through a case study exploring design practices in a project aiming to improve speech technologies for Australian Aboriginal English. We discuss how we integrated culturally appropriate and participatory processes throughout the project. We call for increased support for languages used by Indigenous communities, including contact varieties,  which provide practical economic and socio-cultural benefits, provided that participatory and culturally safe practices are enacted.

\end{abstract}

\maketitle
\section{Introduction} 

Most work on speech and language technologies, a.k.a.\ Natural Language Processing (NLP), focuses on languages which are spoken by majority populations and heavily standardised within one or more countries \cite{bender2009linguistically, sawicki2023state, joshi-etal-2020-state}. 
The field of NLP is thus reproducing the minoritisation of language varieties, and in the process reinforcing barriers of exclusion for diverse language communities. These include most local languages in highly linguistically diverse countries such as Indonesia \cite{aji-etal-2022-one},
as well as various creoles which combine elements of settler-colonial languages elsewhere in the world \cite{lent-etal-2021-language}.
Additionally affected are varieties of national languages spoken by minoritised groups \cite{kantharuban-etal-2023-quantifying}, such as Māori English in New Zealand, 
Multicultural London English in the UK, and 
African-American English in the US. 
Speakers of minoritised varieties must typically switch to a majority variety when using language technologies, and when they do they can experience negative psychological impacts, such as feeling that technology wasn’t designed for them \cite{cunningham2024understanding, mengesha2021don, harrington2022s, wenzel2023can}.

Machine learning-driven NLP is not typically oriented towards participatory processes, driven instead by the valorisation of efficiency \cite{birhane2022values} and fundamentally constrained by tensions between participation and scale \cite{young2024participation}. NLP evaluations emphasise quantitative and decontextualised metrics \cite{hutchinson2022evaluation}, while language dataset projects often reduce participation to crowdsourced platform work. These projects fail to create relationships which emphasise reciprocity and social justice \cite{sloane2022participation,cooper2024fitting}, and risk  perpetuating colonial histories, especially in the Global South, of exploitation by academic institutions, states and corporations \cite{markl2022mind}. However participatory AI projects, on the other hand, are often challenged by a lack of shared understanding of good practices \cite{groves2023going}, and may risk ``participation-washing'' \cite{sloane2022participation}. More generally, projects which aspire to participation often fail to afford decision-making power to participating communities \cite{arnstein1969ladder}, and risk co-optation when delivered through corporate structures \cite{asaro2000transforming, birhane2022power}.

In Australia, where this research takes place, of the 400+ ancestral Indigenous languages that existed before European colonisation \cite[p.\ 56]{bowern2023oxford}, most are either extinct or critically endangered \cite{Australian_Government2020National}. This is due to linguicidal practices of colonialism, which at various times included governments forcibly separating Aboriginal children from their families and coercing Aboriginal pupils to speak English while forbidding them from speaking their languages \cite{Rademaker2018Found}. As a result, most Indigenous people in Australia speak Australian Aboriginal English, making it a strong encoder of ethno-cultural identity \cite{rodriguez2024yarning}.\footnote{Other contact languages in Australia include mixed languages such as Gurindji Kriol, Light Warlpiri and New Tiwi \cite{Meakins13}.}  
This indigenised variety of English integrates phonological, lexico-grammatical and discourse-pragmatic features from English and Aboriginal ancestral languages (see \S\ref{sec:aae}).

Given this context, we take three positions in this paper. First, Indigenous people should be supported to use \emph{whichever of their linguistic varieties that they prefer}, including contact varieties, when using language technologies.
Second, NLP requires not only culturally appropriate outputs, such as models, tasks and evaluations \cite{Hovy2021Importance, hershcovich2022challenges, liu2022not}, but also culturally appropriate processes, outcomes and modes of participation  \cite{noe2024generalized, bird2024centering}. All are currently undervalued in NLP.
Third, language technologists should be equally concerned with contemporary Indigenous needs, including supporting recent and evolving Indigenous language varieties, as they are with needs related to ancestral languages. What, then, are these opportunities, as well as the risks? And how are they distinct from the technology needs of mainstream language communities, or of ancestral Indigenous language communities? 

To address these questions, this paper contributes the first detailed consideration of language technologies to support Australian  Aboriginal English speakers. To our knowledge, this is the first paper to focus on technologies for this language community. We do so through a case study of an ongoing project intended to improve Automatic Speech Recognition (ASR) for Indigenous communities in Australia. In doing so, this paper follows previous NLP literature in considering a specific Indigenous language community and its context   \cite[\eg{},][]{zhang2022can, teodorescu2022cree}, with a  focus on design considerations in order to avoid assuming that mainstream methods and goals are locally applicable  \cite{noe2024generalized, pine2024speech}.
As such, this paper contributes to our understanding of ethical and responsible development of technologies for minoritised communities, as well as to our understanding of how to design and build culturally sensitive technologies while incorporating participatory methods.

The paper begins with an introduction to Australian Aboriginal English, including a description of relevant linguistic and sociolinguistic factors (\S\ref{sec:background}). We also present a brief survey of related work on Indigenous language technologies.
Next, we discuss the opportunities and risks inherent in deploying ASR systems for this community, as well as opportunities and risks in projects building such technologies (\S\ref{sec:risks and opportunities}).
Building on this analysis, we explore strategies to mitigate risk, and facilitate opportunities, through a detailed discussion of the design considerations of a real-world project to design culturally appropriate ASR while integrating participatory methods (\S\ref{sec:asr}).
In doing so, we provide a detailed discussion of some of the considerations that are critical for a project of this sort, including system requirements, data elicitation and transcription, system evaluation, and processes and frameworks for participation, and employment. 
Finally, we discuss  further technological opportunities for supporting speakers of Australian Aboriginal English (\S\ref{sec:opportunities}).

\section{Background and Related Work}
\label{sec:background}

When considering how technologies can support speakers of a specific Indigenous language variety, it is critical to have some basic knowledge of the speech community, the language variety, and their contexts \cite{zhang2022can}.
In this section we introduce Australian Aboriginal English and its context,  followed by a brief summary of prior work relevant to language technologies for Indigenous communities. This includes consideration of both ancestral and contact language varieties, and also of participatory methods in Indigenous language projects.

\subsection{Australian Aboriginal English}
\label{sec:aae}

Australian Aboriginal English is a contact variety of English used by Indigenous people in Australia \cite{louro2021australian}. It is particularly prevalent across southern Australia where ancestral languages are less widely spoken. In these contexts, Australian Aboriginal English is used to strengthen Indigenous ways of being and communicating \cite{eades2012communication}.
Australian Aboriginal English has its {origins} in the product of forced contact between Indigenous people and the English-speaking settler-colonialists. The English spoken by the British was a powerful instrument used to strip Indigenous communities of their ancestral rights \cite[][p.\ 125]{malcolm2000aboriginal}. Australian Aboriginal English’s precursor was a mixed jargon viewed as `broken English' by the British. By the time this jargon had transitioned to a novel pidgin (known as `New South Wales pidgin' \cite{troy1994melaleuka}), it began to travel outside present-day Sydney and across New South Wales. With the advent of the pastoral industry, the pidgin also spread north into Queensland and the Northern Territory, giving rise to Queensland Pidgin English and the Northern Australian creoles, including Roper River Kriol \cite[][p.\ 371]{meakins2014language}. 

Australian  Aboriginal English has {linguistic characteristics}, including pronunciation, lexis, grammar, and discourse-pragmatics, that distinguish it from mainstream English \cite{eades2012communication}. Pronunciation features include: frequent /h/-dropping and /h/-addition; metathesis of `k' and `s' in the word `ask', which is pronounced `aks'; and default reduction of the `ing' verbal suffix to `in'. Grammatical features include  double negatives as in `never done nothing'; and  double subjects as in `My niece, she did it.' Lexical features include lexical transfer such as \emph{blackfella}, meaning `black person/fellow', and  borrowings from ancestral languages, such as \emph{dardy} `good/good looking' from Nyungar. 

Experts remain ambivalent as to whether Australian Aboriginal English represents a single ethnolect, or whether different varieties of Aboriginal English, or `Aboriginal Englishes', exist across Australia \cite{dickson2019aboriginal}.  
Australian Aboriginal English is frequently placed on a continuum that ranges from `lighter' {acrolectal varieties} (closer to standardised Australian English) to `heavier' {basilectal varieties} which are closer to Kriol, spoken across northern Australia \cite{schultze2013kriol}.
The basilectal varieties are mostly unintelligible to standardised English speakers, but the acrolectal varieties are superficially similar to Australian English.

Because Australian Aboriginal English is English based, it is sometimes {socially stigmatised}, erroneously classified by laypeople as an incorrect version of English. As a result, L1 users of Australian Aboriginal English are often discriminated against in schooling and legal settings \cite{harkins1994bridging, eades2012social}, perpetuating invisibility of the variety and leading to the false perception that its speakers are uneducated and rude \cite{eades2013aboriginal}.
Stigmatisation of Australian Aboriginal English is evident in the comment below, which was offered in response to an online article on the role of Australian Aboriginal English in Indigenous health
communication \cite{rodriguez2021yarns}.
\begin{quote}
With Aboriginal children all going to school,
and nearly all of them being taught in English, 
does it still serve them to support the continued use of Pidgin English?
\end{quote}
The commenter's deliberate use of `Pidgin English' reflects a deficit perspective that undermines the value of
Australian Aboriginal English. Rather than embracing 
Australian Aboriginal English as 
a linguistic variety many Indigenous people use daily in Australia, this perspective portrays racialised peoples' language varieties as
`lacking' vis-à-vis mainstream English \cite[p.\ e214]{hudley2020toward}.

These negative attitudes are exacerbated because Australian Aboriginal English is {primarily 
oral}, with no nationally adopted guidelines as to how it should be
written. Additionally, the orthographic representation of Australian Aboriginal English varies
across locales, reflecting the local linguistic biographies and practices of distinct
Indigenous groups. For example, the name of the Traditional Owners of the region
comprising the southwest of Western Australia,
as well as their language, is variably spelled ‘Nyungar’,
‘Noongar’, or ‘Nyoongar’ depending on community members' preferences.\footnote{\glenys{} is Nyungar so we prefer this spelling.} Rather than a weakness, the lack of standardising
orthographies for Australian Aboriginal English reflects Indigenous people’s prerogative to
differentiate---rather than homogenise---how Australian Aboriginal English is used across
Indigenous Country.

Indigenous people tend to find the use of direct questions awkward or offensive \cite{rodriguez2024hearing},\footnote{With some exceptions, including teaching contexts where instructors asking questions is heavily routinised \cite{moses2008questions}.}  and these {pragmatic considerations} continue in Australian Aboriginal English. 
Direct questions are also likely to elicit misleading information; Indigenous people
have been known to answer ‘yes’ or agree with non-Indigenous people even if the
question is not fully understood—a phenomenon known as `gratuitous concurrence'
\cite[p.\ 137]{liberman1985}. To avoid these limitations, researchers have used `yarning',
an Indigenous cultural form of storytelling and conversation, to lead unstructured
data collection with Indigenous communities who speak Australian Aboriginal English  \cite{Bessarab10,rodriguez2021working}.

\subsection{Technologies for ancestral Indigenous languages}

The NLP literature for Indigenous languages tends to focus on ancestral languages, \textit{i.e.}, Indigenous languages which predate colonisation.\footnote{Others call these languages `traditional' Indigenous languages, \eg{}, \cite{eades2013aboriginal}. The term `ancestral' is sometimes used by others with a different meaning: a language from which modern languages descend.} 
Given the considerable overlap between endangered languages and Indigenous languages \cite{Bromham2022}---a consequence of colonisation---NLP literature tends to focus on application domains which support language revitalisation and preservation efforts, including assisting language learners or teachers \cite{meighan2021decolonizing, meighan2022indigenous, galla2009indigenous}. This literature discusses technologies such as quiz generation, automated language assessment, language learning platforms, computer assisted translation, predictive text, and readability classification \cite[\eg{}][]{zhang2022can,teodorescu2022cree,carew2015getting,mager2024proceedings}. \citet{bird2024centering} pose open questions regarding distinct needs of learners of oral languages without the technology working through a written form. 

Language preservation or documentation is frequently cited as a goal \cite{le2023leveraging, prud2021automatic}, and is discussed variously as a means towards revitalisation, or as an end in itself.
Recognising that ancestral language communities can be minoritised in multiple ways, \citet{carew2015getting} discuss the role of language technologies in promoting digital inclusion. Other motivations encompass the importance of language technologies in cultural revitalisation projects, \eg, helping to learn about traditional place names, stories, songs and kinship relations.

\subsection{Technologies for contact language varieties}

The NLP literature has to-date had marginal concern with supporting Indigenous contact language communities, such as Hawaiian Pidgin, Australian Kriol, Tok Pisin, Chavacano, \etc{}. For example, Nigerian Pidgin has over 100 million speakers \cite{lin2024modeling}, yet is poorly supported by technology. This seems to be due to a combination of reasons, including the  stigmatisation of contact languages \cite{siegel1999creoles, lent2024creoleval} and a belief that ancestral languages are more important in terms of global heritage.
However Indigenous peoples are not museum artifacts or relics of an `ancient' era 
\cite{gaby2020toward, Hill02};
their cultures and societies are dynamic and evolving.

To understand the needs of Indigenous contact language communities, we first draw on the NLP literature for Creole languages. \citeauthor{lent2022creole}'s survey of six Creole communities found different levels of desire for language technologies \cite{lent2022creole}. They note that these languages often lack recognised status, are primarily oral with variable  orthography and writing conventions, and are often socially stigmatised.
This points to the importance of contact varieties in shaping contemporary social identities. If technology does not support contemporary forms of Indigenous expression, then it reproduces and reinforces minoritisation of Indigenous culture and identity.
Contact varieties are often  vehicular, used to communicate with other communities, including participation in national and regional economies, and with civic services \cite{bird2022local}. If language technologies are to support economic and civic participation, 
then they need to support vehicular varieties.

Since contact varieties are primarily oral, prior work has identified {speech technologies} as the most wanted by both experts and communities \cite{lent2022creole,le2023leveraging}. Robust Automated Speech Recognition (ASR) that can recognise Indigenous pronunciations and loan words is an important foundational technology for a broad range of applications, from voice search to automated transcription and interaction with digital assistants \cite{aksenova-etal-2021-might}. Design of ASR for contact varieties should be mindful of the social dimensions of orthography, in particular of variation from standardised written forms.

For some contact varieties, orthographic norms may have colonial associations, and can be alienating for speakers \cite{markl2023automatic}. For this reason, and due to their primary orality, consultation with experts and communities is required to ascertain the need for text-based technologies \cite{lent2022creole, leferrand2022learning, bird2022local}. When needed, {writing tools} supporting local scripts, fonts, and keyboards would generally be the most useful text-based technologies \cite{vanesch2019writing}. 
The lack of standardised writing conventions for many contact varieties may present challenges for the utility of  technologies such as spell checkers and next word predictors.

Language technologies are increasingly multilingual, and a critical component of many systems is {Language Identification} (LangId).
Since most Indigenous communities are multilingual, robust LangId and code-switching/mixing detection are needed to support local multilingual technologies \cite{nguyen2013word}.

\subsection{Culture, and participatory methods, in language research}

Recently, the NLP community has demonstrated increasing interest in the topic of culture, 
with, \eg{}, three recent workshops on Cross-cultural Considerations in NLP \cite{c3nlp-2023-cross,c3nlp-1-2024,c3nlp-ws-2025-1}. This interest has been spurred by the capabilities of Large Language Models (LLMs) with regards to linguistic form, while also demonstrating lack of cultural nuance \cite{hershcovich2022challenges,naous.etal2024beer, li2024culturellm}. Threads of research have focused on measuring cultural biases \cite{naous.etal2024beer,Tao2024Cultural} and expression of cultural values by LLMs   \cite{Arora2022Probing,Durmus2023Towards,Ramezani2023Knowledge}.

It is critical that the NLP community goes further than developing technological fixes in data and models, 
to consider design questions and project processes. NLP research using Indigenous language data often fails to engage with the cultural differences between technologists and the communities they aspire to support \cite{bird2024centering}. Recent community-led projects such as Masakhane \cite{nekoto2020participatory} and the Papa Reo project \cite{jones2023kia} provide alternative models for the NLP community, by drawing on principles of participatory design to embed the cultural knowledge of language communities throughout NLP research and development processes.
Rather than developing technologies \textit{for} Indigenous contact varieties in isolation of communities speaking those varieties, the primary focus of such projects must be on developing language technologies \textit{with} Indigenous language communities. This requires research teams, and the broader NLP community, to consider the position of NLP researchers relative to community partners, reflecting not only on performance disparities of technologies across languages or cultural biases, but also on
the composition of research and development teams, as the cultural assumptions and norms of such teams influence the research we conduct \cite{forsythe2001studying}.

\subsection{Projects with Indigenous communities}
\label{sec:projects_with_communities}

The literature emphasises that working with data of Indigenous language varieties raises ethical considerations. These also apply to contact varieties such as Australian Aboriginal English, and are critical to keep in mind when considering opportunities and risks. Four questions that recur in the literature include:

\paragraph{Q1.\ What is the  context?}
Before working with Indigenous varieties, researchers must reflect on the harmful impact of past colonial research projects on Indigenous communities and seek to understand the historical and sociolinguistic context of partner communities \cite{smith2021decolonizing, schwartz2022primum, dougruoz2022language, bird2020decolonising, zhang2022can, noe2024generalized, leferrand2022learning}.

\paragraph{Q2.\ What relationships will be maintained?}
Relationships are fundamental to Indigenous identities, communities, and research \cite{smith2021decolonizing, wilson2020research}, and this holds for research on contact varieties. Technologists should be cognisant that relationships with partners and participants may, and in some cases should, persist beyond the scope of the project \cite{cooper2024s}.

\paragraph{Q3.\ Who will control the project, and its outputs?}
In both ancestral and contact variety-speaking communities, 
Indigenous people are less likely to instigate language technology projects, enumerate technological goals, or design/control project budgets, and it is important to both acknowledge and address these power differentials.
Using language data ethically is the lowest threshold for language projects \cite{cameron1993ethics}, and involves considerations not only of IP (Intellectual Property) but also ICIP (Indigenous Cultural and Intellectual Property) \cite{janke2019true}.

\paragraph{Q4.\ Is the technology appropriate for the community?}
Indigenous language varieties are typically differentiated functionally within a multilingual community, and technologies may be more useful for some  than others \cite{dougruoz2022language, bird2022local, lent2022creole}. Technologists should understand the technology needs of Indigenous communities, which may be distinct from those of non-Indigenous users \cite{liu2022not, cooper2024s}.

\section{Opportunities and Risks for ASR for Australian Aboriginal English}
\label{sec:risks and opportunities}

We now consider Automated Speech Recognition (ASR) systems for Australian Aboriginal English, acknowledging that this focus on converting spoken language to written language  itself risks reproducing beliefs about the primary importance of written language vis-à-vis spoken \cite{chrupala-2023-putting}. 
With the contexts and questions of the previous section in mind, we consider potential opportunities and risks, informed by the literatures on ASR, Human-computer Interaction, and Indigenous language technologies. We consider both deployed ASR technologies as well as the projects which build ASR technologies; our ultimate goal is to build inclusive technologies using inclusive methods.

\subsection{Deployment Opportunities}
\label{sec:system opportunities}

ASR supports many application domains \cite{aksenova-etal-2021-might}, and improving ASR for Australian Aboriginal English could provide multiple specific benefits to First Nations people in Australia, their communities, and non-Indigenous people who are encountering Australian Aboriginal English. We attempt to distinguish those possible benefits with as much specificity as possible, and in contrast to prior work we adopt a human-centered framing rather than a technology-centered one.

\paragraph{Opportunity 1: Improved dictation for Australian Aboriginal English speakers.} Dictation typically involves a single speaker consciously speaking directly to a device, and is useful in a variety of contexts in which speaking is preferred to typing. These include messaging while driving, speakers with physical or vision disabilities, speakers avoiding small keyboards on mobile devices, and speakers with dyslexia. The speaker often reviews the text output for accuracy. Since they are conscious of the interaction, some speakers may modify their speech in an attempt to improve the fidelity of the transcription, such as minimising use of informal registers, obscure or complex words,  and minoritised varieties \cite{mengesha2021don,holt2024perceptions}. 

We note that not all dictation applications are likely to benefit Australian Aboriginal English speakers. For example, ASR is sometimes used in \emph{dictation practice} by second-language learners, however Australian Aboriginal English is not formally taught or studied as a second-language.

\paragraph{Opportunity 2: Improved transcriptions for readers of personal messages.} ASR can be initiated by those within the speaker's community or social network, for example, users of a voicemail service providing automated transcriptions. This can be useful in a variety of contexts in which  reading is preferred to listening, including in quiet environments in which silence is preferred, in noisy environments in which hearing is difficult, when  privacy is desirable in public environments, and for people with hearing disabilities. An Australian Aboriginal English speaker is often unaware when their voice is being transcribed, and so is unlikely to modify their voice. The reader might or might not listen to the voice recording to verify fidelity of the transcription. The reader may be able to accommodate and even correct some ASR errors, based on their familiarity with the speaker and/or context, even without listening to the recording.

\paragraph{Opportunity 3: Improved transcriptions for consumers of platform content.} When speech and video platforms host content, they often provide closed captions. Such content can be unscripted (typical for Australian Aboriginal English on YouTube, for example, see Appendix~B) or scripted (common in TV shows, movies, podcasts and vodcasts, etc---for an analysis of scripted Australian Aboriginal English see \cite{bednarek2021australian}). These are generated offline and stored as metadata alongside the Australian Aboriginal English voice recording. The platforms may have some access controls, but in the general case both the recordings and the transcriptions may be accessible to potentially anyone online, including listeners and readers with no prior experience with Australian Aboriginal English. Listening to the recordings might not help readers without familiarity with Australian Aboriginal English to correct some ASR errors.

Due to systemic societal biases, some applications of closed captioning may encounter relatively little Australian Aboriginal English compared to Australian English. The automated transcription of Australian parliamentary proceedings \cite{gollan2005cross}, for example, is one use case for Australian Aboriginal English ASR that is impacted both by the under-representation of Indigenous people in Australian parliaments and by the social stigmatisation of speaking Australian Aboriginal English in formal contexts.

\paragraph{Opportunity 4: Improved human-machine interactions for Australian Aboriginal English speakers.} ASR is increasingly incorporated into a wide range of computational systems used by Australian Aboriginal English speakers. These include automated voice assistants, automated or semi-automated call centres including for critical banking and government services, web search using voice queries, and audio/video retrieval that indexes speech under transcribed terms. ({Voice Query-By-Example technologies \cite{parada2009query} do not require ASR but constitute a minority of audio retrieval systems.})  In many cases, the transcription might not be seen by any human, including the Australian Aboriginal English speaker. Instead, the transcription is typically the input to a Natural Language Understanding component, which outputs some form of semantic representation in forms such as a task frame, a search index entry,  or an expanded Information Retrieval query \cite{efthimiadis1996query, jurafsky2008speech}. More accurate transcriptions can lead to more relevant search results, more efficient interactions, and fewer abandoned tasks.

\subsection{Deployment Risks}
\label{sec:system risks}

\citeauthor{schwartz2022primum} has argued that the primary ethical responsibility when building language technologies for Indigenous communities is to not do any harm \cite{schwartz2022primum}. Although many taxonomies of risks and negative impacts of technologies exist, our goal here is be a specific and situated as possible. Hence we do not attempt to map the identified risks to prior taxonomies which often aim at universalism and generality \cite{berman2024troubling}. 

\paragraph{Risk 1: Inappropriate for the contexts of Australian Aboriginal English.} Researchers should not assume that minoritised language communities have the same wants and needs as mainstream language communities \cite{bird2022local}, nor that multilingual speech communities require the same technologies for all languages \cite{lent2022creole,  bird2024centering}. For example, would some Australian Aboriginal English speakers prefer to switch into Standardised Australian English when interacting with machines, if it results in greater task efficiency? What about when dictating messages to family and friends? Conversely, would they prefer more support for speaking Australian Aboriginal English in contexts which are currently not well supported? There is a shortage of research on the real-world  needs of Australian Aboriginal English speakers. Building systems that are neither needed or wanted is not only inefficient---it also undermines the rights of minoritised communities to determine which technologies are built for them. 

\paragraph{Risk 2: Lack of sociolinguistic ecological validity.} Like all language varieties, Australian Aboriginal English is not homogeneous, but exhibits regional variation \cite{dickson2019aboriginal}. Furthermore, and notably, Australian Aboriginal English includes both acrolectal and basilectal varieties (\S\ref{sec:aae}). If the training data for an ASR model does not represent sufficient variation, then the model will fail to recognise the breadth of forms of  Australian Aboriginal English. If the training data collection protocols are not designed to be culturally safe, then speakers may  switch into Standardised Australian English, reducing the validity of the data.

\paragraph{Risk 3: Psychological harms for Australian Aboriginal English users when dictating or interacting with systems.} Prior research has found that when ASR systems make errors, some members of minoritised language communities feel that the computer `doesn't like' how they are speaking \cite{brewer2023envisioning}, or that the technology isn't built for people like them \cite{mengesha2021don}. Whereas speakers of mainstream language varieties may blame the ASR system for poor performance, speakers of minoritised communities may blame themselves, and experience a drop in self esteem or an increase in self consciousness about their identity \cite{wenzel2023can}. They may feel that their language variety is being judged as illegitimate \cite{cunningham2024understanding}. If ASR fails to meet expectations for speakers of Australian Aboriginal English, all of these impacts may be reproduced. Feelings of shame or embarrassment might also arise if speakers are self-conscious about their way of speaking being transcribed and re-contextualised into contexts where mainstream written English is the norm.

\paragraph{Risk 4: Torquing of Australian Aboriginal English sociolinguistic identities.} ASR systems function as classifiers not just of speech, but also, by extension, of language varieties and sociolinguistic identities. As such, ASR systems can be examined through Bowker and Star's sociological lens of torque and residuals \cite{bowker2000sorting}.
`Residuals' refers to the `other' categories that are relevant to the individual but missing from the classificatory system, while `torque' refers to the painful twisting and bending that occurs when an individual's data is forced into an inappropriate classificatory system, such as forcing individuals into racial categories in Apartheid South Africa. Previous work has explored how gender classification technologies produce torque and othering for trans people \cite{scheuerman2019computers}, and how medical and biometric IT projects can torque minoritised groups \cite{avlona2024torquing, singh2017margins}. We posit that a similar form of torque exists for users of non-standardised language varieties, when ASR systems systematically fail to recognise spoken forms salient to social identities, bending and twisting them into the written forms of  mainstream dictionaries of colonising cultures.

\subsection{Project Opportunities}
\label{sec:project opportunities}

We take the position that language technology projects provide the potential for a  range of opportunities, especially when the project is conducted not just \emph{for} the community but also \emph{with} the community \cite{cameron1993ethics}.

\paragraph{Opportunity 5: Project leads to greater non-Indigenous understanding of Indigenous contexts and histories.} Projects involving both Indigenous and non-Indigenous participants create opportunities for `two way learning', a phrase used in Australia to refer to Indigenous and non-Indigenous people coming together to learn from each other in a space that is safe for those with diverse perspectives and ways of knowing \cite[p.\ xx]{purdie2012two}.    

\paragraph{Opportunity 6: Project work produces opportunities for Indigenous community.} Data collection projects can provide opportunities for what \citet{sloane2022participation} call `participation as work', with equitable compensation. There are large economic disparities in Australia, with median incomes for Indigenous people around 23\% lower than for non-Indigenous \cite{Australian_Institute_Health_welfare_2023Income}, and Indigenous people have higher rates of unemployment for all education levels \cite{Australian-Institute-of-Health-and-Welfare2023Employment}. If equitable compensation and non-extractive processes are successfully adopted, data work can create economic opportunities for community, as well as opportunities for meaningful cultural engagement \cite{bird2022local}.

\paragraph{Opportunity 7: Project builds capabilities in community.} 
Building Indigenous capabilities on technology projects has the opportunity of addressing the current digital divide in skills \cite{carew2015getting}. When projects effectively build community capacity, participants can draw on their developing technical skills---such as data collection and management capabilities---to advocate for their communities in future projects and on technology and policy issues \cite{cooper2022systematic}. To achieve these benefits, meaningful opportunities for skill development need to be incorporated into the project, focusing on capabilities that will remain valuable to the community beyond the project's completion. 

\paragraph{Opportunity 8: Project increases Indigenous self-determination.} Providing opportunities for Indigenous control over project outputs, for example through data governance. Design decisions can have not only consequential benefits (\eg, avoiding data set misuse by third parties in ways that do not accord with Indigenous aspirations), but also constitute a form of procedural justice through greater self-determination.

\subsection{Project Risks}
\label{sec:project risks}

Developing inclusive ASR \emph{inclusively} requires more than addressing model biases \cite{feng2024towards}--- it also requires researchers to integrate participatory practices throughout the project. 
As such, we take the position that, when developing language technologies, risks need to be considered using a wide-angle lens that takes into account not just the deployment scenarios but also the project processes used when developing the technology. For projects with Indigenous communities, `it's how do you things that matters' \cite{cooper2024s}, and the ends should not be used to justify the means.

\paragraph{Risk 5: Project harms relationships with community.} The literature on building language technologies for Indigenous communities emphasises the foundational importance of relationships with the community \cite{zhang2022can, cooper2024s, bird2024centering, bird2020decolonising, bettinson2021designing}. If a project damages community trust and harms relationships, not only will that project become endangered, but future projects with the community may also fail due to lack of community support.   

\paragraph{Risk 6: Project fails to partner with community where appropriate.} A lack of community participation, generally, can not only lead to poorer data and models, but can also miss opportunities to address social injustices \cite{sloane2022participation}. Failure to partner when appropriate with Indigenous communities, specifically, can reproduce histories of ignoring or disrespecting Indigenous sovereignty.  Calls for collaborative and participatory research recognise community rights to be meaningfully included in language research that involves them \cite{carew2015getting, cameron1993ethics}.  This is emphasised by \textsc{unesco} in the Los Pinos Declaration calling for a decade of action on Indigenous languages: `Nothing about us without us' \cite{unesco2020pinos}.

\paragraph{Risk 7: Project uses culturally unsafe methods to collect data.} Recording one's own voice can be an uncomfortable experience for many people, and especially so for members of minoritised communities encountering researchers from universities or technology companies using mainstream English. Unless care is taken, prompts might not be culturally relevant, stray into territories of secret traditional knowledge, elicit personal narratives of trauma, or trigger feelings that their culture is not important and valid. 
Data work may also risk being meaningless to workers' social or cultural contexts, for example, transcribing long recordings of a non-standardised variety into a mainstream script can be tedious and unengaging work \cite{bird2022local,bird2021sparse}.  

\paragraph{Risk 8: Project does not respect Indigenous data sovereignty.} Failure to adopt appropriate Indigenous data governance protocols can violate the principles of Indigenous people controlling Indigenous data \cite{carroll2020care,Maiam_nayri_Wingara2018, Te_Mana_Raraunga2016}.  This is an important aspect of generally recognising Indigenous sovereignty, which was never ceded in what we now know as Australia. 

\paragraph{Risk 9: Project does not acknowledge Indigenous knowledge and expertise.} There are many ways in which Indigenous contributions may fail to be acknowledged. Within technology research, this includes not citing or including as co-authors Indigenous experts whose ways of working fall outside Western academic norms \cite{faulkhead2023indigenous}.  

\section{Designing ASR for Australian Aboriginal English}
\label{sec:asr}

We now discuss a real-world partnership between \google{} and \uwa{} which aims to improve ASR for speakers of Australian Aboriginal English.
We focus on the design considerations, and specifically on our efforts to relate the project goals and processes to both the sociolinguistic context (\S\ref{sec:aae}) and the technological opportunities (\S\ref{sec:system opportunities}). We discuss not just technical questions, but the design of the project itself, including how we attempt to mitigate the system and project risks (\S\ref{sec:system risks}, \S\ref{sec:project risks}) while providing societal and economic opportunities (\S\ref{sec:project opportunities}), \textit{e.g.}, by integrating participatory processes throughout the project lifecycle.

When building technologies for Indigenous speech communities---of either ancestral or contact varieties---it is important to consider the ways in which projects face social and technical questions distinct from those that involve mainstream languages \cite[see, \eg{},][]{pine2024speech, bird-2020-sparse, carew2015getting, noe2024generalized, lent2022creole, leferrand2022learning,taylor2020}.
For example, ASR development pipelines typically make assumptions about speech communities and language use, including the existence of `gold standard' spoken and written forms \cite{markl2023automatic}.
These assumptions break down in the case of Australian Aboriginal English, for which the written form is not standardised, and so community participation in the design process is a prerequisite for building technologies that serve the community.

\subsection{System desiderata}

ASR should support the use of Australian Aboriginal English in all application horizontals which incorporate ASR systems. These include: 1) dictation of  messages and documents; 2) voice queries to search engines; 3) voice commands to virtual assistants; and 4) automated transcriptions of  voicemail messages, meetings, podcasts, and video content \cite{aksenova-etal-2021-might}. It is important to distinguish applications in which the transcripts will be processed by computers (Opportunity 4 in \S\ref{sec:risks and opportunities}, \eg{},  commands to virtual assistants)  from those where the transcripts will be read by humans (Opportunities 1--3, \eg{}, voicemail transcriptions). In the first case, post-processing converts the transcripts to a structured or semi-structured representation, such as an expanded Information Retrieval query or a task frame \cite{jurafsky2008speech}. This post-processing involves increasing abstraction and standardisation, mapping diverse user inputs into more computationally tractable forms. In the case of transcripts for humans, however, standardisation of linguistic forms is no longer a computational requirement.\footnote{The requirement for readability by non-linguists also precludes transcription into phonetic or phonemic representations, as has been done for some other work such as \cite{michaud2018integrating}.}  Instead, standardisation of transcripts minimises the diverse ways in which social meanings are conveyed through linguistic choices. In the process, standardisation reproduces the minoritisation of non-mainstream sociolinguistic identities  \cite{markl2023automatic}.  This is especially relevant for informal registers, such as those used in personal messages or social media, in which expressions of social identities are more prevalent and varied.

\subsection{Elicitation of speech data}


The collection protocols to secure the speech data needed for this project were designed by the authors with advice from the Indigenous Advisory Committee (\S\ref{sec:project_before}). 
The project also applies participatory research methods discussed further in \S\ref{sec:considerations} and previously used by \celeste{} and \glenys{} \cite{rodriguez2024yarning}. However, the data to be collected for this project will be used by \google{} and it is expected that the corpus will be open access, in accordance to FAIR and CARE principles \cite{wilkinson2016fair, carroll2020care}, subject to further input from the Indigenous Advisory Committee. For this reason, the project explicitly excludes sharing of taboo cultural materials and sensitive information that is potentially harmful for Indigenous people and communities. To ensure cultural safety (cf.\ Risk 7), we designed a program of careful elicitation by using culturally relevant and sensitive prompts in consultation with the Indigenous Advisory Committee for this project, and ultimately designed by an Indigenous-owned company which specialises in producing educational resources using an anti-racism lens (Appendix~A, cf.\ Risk 4).

The design of the recording process enables voice data providers to record their voices in their own homes, which is a more comfortable and culturally-sensitive experience than visiting a university or a corporate office. Participant payment processes were designed with both safety and convenience in mind (Opportunity 6).

Once collected, all data will be screened by a team of Indigenous RAs, led by \glenys{}, who will vet the audio files and transcriptions for cultural sensitivity. Any culturally unsafe material (\eg, stories of specific people, places, Indigenous knowledges, trauma, and loss) will be erased from the corpus before sharing the data with others.

\subsection{Transcription of speech data}

The introduction of writing by settler-colonialists has significantly impacted Indigenous populations \cite{ong2013orality, franchetto2008war}. In Australian Aboriginal English, there are two major questions regarding orthographic practices. First, how are loan words from ancestral languages, including place names and Country names (\eg{} `Nyungar'), written in Latin-based script? Second, how do Indigenous people in Australia deliberately adopt their own spelling conventions, \eg{},  `Blak'. 
These are not purely technical questions, but also sociopolitical \cite{franchetto2008war, markl2023automatic,bucholtz2000politics} (cf.\ Risk 4).


To co-develop transcription guidelines with community members, we required examples of public Australian Aboriginal English to refer to. For this, we used a corpus of over 100 hours of YouTube videos that are primarily in Australian Aboriginal English (see Appendix~B).  An Indigenous team, led by \celeste{}, developed the corpus and used their expertise with Australian Aboriginal English to validate the variety spoken, ensuring that speakers were only coded as Australian Aboriginal English if they used at least two linguistic features that distinguish this variety from Standardised Australian English. We developed an initial version of variety-specific transcription guidelines using examples from this corpus. Example guidelines include:
\begin{enumerate}
     \item borrowings from Pidgin and ancestral languages should be transcribed verbatim, with an English translation included;
    \item variety-specific utterance tags should be included, \eg{}, `unna' and `inni'.
\end{enumerate}
The transcription guidelines are a living document, which will grow---with input from the community---as novel classes of examples are observed.

\subsection{Evaluation considerations}

\citet{dougruoz2022language} have observed that NLP metrics are usually designed with high-resource scenarios in mind.
In the case of ASR, the most common evaluation metric is Word Error Rate (WER), which does not actually compare words but rather spelled forms \cite{karita-etal-2023-lenient}. 
It does so in binary fashion: small differences between predictions and reference transcriptions (\eg{}, `na\"ive' vs `naive') are penalised just as much as large ones. As such, WER is less useful as a measure of system utility in various contexts, including with language varieties without standardised written forms  \cite{ali2015multi,nigmatulina-etal-2020-asr}, and varieties with multiple standard written forms \cite{kummervold2024whispering, karita-etal-2023-lenient}. In the case of Australian Aboriginal English, WER is not robust to culturally acceptable variation in borrowings from ancestral languages (\eg{}, `Gadigal' vs `Cadigal'\footnote{The Country and people of the Sydney region.}) or indigenised spelling variants of English words (\eg{}, `Blak' vs `Black').\footnote{Character Error Rate (CER), another common metric, does not adequately resolve the issue. CER can avoid the overpenalisation of variants accepted by different groups in the community, \eg, only penalising lightly a prediction of `Cadigal' when the reference transcription is `Gadigal' (both refer to the same First Nations Country). However CER is too lenient a metric in cases where the ASR system confuses the spoken word for another word which has a  very different meaning or referent, \eg, predicting `Kukatj' when then reference transcription is `Kukatja' (the names of two unrelated  Aboriginal language varieties with ancestral lands thousands of kilometres apart \cite{Austlang2025}). CER would judge such an error as equally bad as the `Cadigal' vs `Gadigal' case.}  

With a lack of standardised orthography for Australian Aboriginal English, it is important to engage with communities, to centre their perspectives, rather than relying solely on automated metrics such as WER \cite{cunningham2024understanding,markl2023automatic, bird-2020-sparse}. As such, the design of the evaluation stage includes interviewing Aboriginal community members, to understand their perspectives on how important fidelity to references is  compared to other considerations.
An Australian Aboriginal English-specific mapping of commonly accepted spelling variations will be incorporated into WER calculations and enable quantitative measures which provide increased validity with community practices \cite{ali2017werd, nigmatulina-etal-2020-asr} (cf.\ `lexeme error rate' of \cite{bird2021sparse}).


\subsection{Designing the project}
\label{sec:project_before}

The project is a partnership between \google{}  and \uwa{} (UWA). \google{}  is a technology company, many of whose products incorporate ASR technologies. \uwa{}  is an Australian University whose Linguistics department has both expertise in Australian Aboriginal English and existing relationships with Aboriginal communities, including the  Nyungar community. We recognise that the legacy of unethical and settler-colonial research has contributed to distrust of research in many Indigenous communities \cite{smith2021decolonizing}. This legacy, along with a distrust of technology companies, can jeopardise even well intentioned projects. \google{} found that seeking out partnership with (socio)linguists at \uwa{} with existing relationships, cultural competencies, and sociolinguistic expertise, was critical, and preferable to attempting to build them from scratch \cite{dougruoz2022language,cooper2024s}. With regards to control of the project and its outputs (Q3 of \S\ref{sec:projects_with_communities}), the project is funded by \google{}, who set the technological goal of the project: improving ASR for First Nations people in Australia. Google will own any ASR models that they develop using the voice recordings, however UWA will retain ownership of the recordings and serve as stewards of the collection for its appropriate use, by setting up culturally appropriate data governance. We strengthened accountability through explicit incorporation of ICIP considerations in our research partnership agreement (\S\ref{sec:considerations}) (cf.\ Risk 8).

Before funding was secured for the project, Aboriginal scholars (with expertise in education, applied linguistics, and language revitalisation) were approached and asked to participate as members of a committee advising researchers on the project. Because the research involves Indigenous participants, letters of support from advisors Professor Clint Bracknell and Mx Sharon Davis, and from \textsc{aiatsis} (Australian Institute of Aboriginal and Torres Strait Islander Studies), were included in a submission to \uwa{}'s IRB. Both \uwa{}'s IRB and \google{}'s  privacy and legal teams reviewed the consent forms that would be used with Aboriginal people creating voice recordings. 
In addition, we sought and obtained the support and expertise of the Language Data Commons of Australia (LDaCA),\footnote{\url{https://www.ldaca.edu.au/}} as the research involves creating a new dataset of Indigenous language materials. LDaCA advised on culturally appropriate data governance and access frameworks \cite{foley2024access}. 

The collection of Australian Aboriginal English voice recordings for the purposes of this project raises Indigenous Cultural and Intellectual Property (ICIP) issues. As a team, the academic authors at \uwa{} negotiated with \google{} the inclusion of an ICIP clause in the research agreement signed in November 2023. This clause reads as follows:

\begin{quote}

The parties acknowledge that Australian protocols may be relevant to ICIP in relation to the Research and will work collaboratively during the Research Term to agree the extent to which Australian protocols, including but not limited to the Australian Council for the Arts Protocols for using First Nations Cultural and Intellectual Property in the Arts (2019), should apply to the Research.

\end{quote}

Most ancestral languages are extinct or endangered, and Australian Aboriginal English is   spoken  by around 80\% of First Nations people in Australia.
This project focuses on both Australian Aboriginal English varieties spoken in southwest Western Australia, where \celeste{} and \glenys{} live and work, and varieties of Australian Aboriginal English spoken in the Pilbara (northern Western Australia), Queensland and Alice Springs (Northern Territory) (cf.\ Risk 2).

Project members took time to reflect on power dynamics, both within the project and in its context \cite{noe2024generalized, blodgett-etal-2020-language} (Opportunity 5). We shared and discussed critiques of extractive data practices. We discussed social injustices, including barriers to opportunity for Indigenous people in Australia, and stigmatisation of Australian Aboriginal English. 

It was important from the outset to be clear about which Indigenous people and organisations would participate in the project, and how their participation would be facilitated \cite{schwartz2022primum, zhang2022can, lent2022creole, bird2020decolonising, flavelle2023strengthening, cooper2024s} (cf.\ Risk 6). In doing so, it was important to be cognisant of the barriers to technology and research careers facing Indigenous participants, to recognise qualifications and authority beyond that of Western institutions, and to recognise contributors as co-authors of research artefacts (\citealp[p.\ 219]{woods2023something}; \citealp[pp.\ 28--29]{meakins2018understanding}) (cf.\ Risk 9). 
While Indigenous participation is critical to the project, we acknowledge that this does not guarantee that the project will be equitable \cite{noe2024generalized, sloane2022participation}. 
We also recognise that the value given to participation and relationships is asymmetric between various project participants. 
We were also mindful not to force control onto community partners, who are often under-resourced and overburdened \cite{cameron1993ethics, corbett2023power}. Establishing the Indigenous Advisory Committee (see below) was an important step in helping to negotiate these tensions. 
We also incorporated aspects of community knowledge and practices into the design process \cite{Theophilus2013Toward} (cf.\ Risk 6), including hiring Goorlil Consulting, a Canberra-based consulting firm owned and run by Indigenous experts, to curate and design  culturally appropriate  visual prompts for the data collection (cf.\ Risk~7). 

An Indigenous Advisory Committee was  established to provide guidance throughout the project. The advisors include Professor Clint Bracknell and Mx Sharon Davis. Bracknell is a respected Nyungar academic with a wealth of expertise in projects examining how Indigenous data should be collected and archived for maximum cultural safety. Davis is a renowned education leader from both Bardi and Kija peoples of the Kimberley in Western Australia. Davis has expertise in, and is a champion of, Australian Aboriginal English. 
Collaborating with Indigenous partners who have expertise in working with community using cultural safety frameworks helps us mitigate risks involving potential harm to Indigenous people and communities (cf.\ Risk~5).
Consultation with the Advisory Committee takes place quarterly. The Advisory Committee meetings, conducted online due to members living in disparate locales across Australia, involve project updates delivered by Rodríguez Louro, followed by Q\&A on project progress and outcomes. The Advisory Committee’s role is to provide critical feedback on project conduct and progress. For example, our first meeting involved discussing culturally-safe ways to collect the project data, while additional meetings discussed how to share the collected data, and how to ensure that the bulk of the research assistants for this project are Indigenous. 

Indigenous associates participate in the project, not just as mere providers of voice data, but as partners in the design of elicitation materials and transcription guidelines, management of the voice data collection, and validation and annotation of the collected materials (Opportunity 7).
All of the research assistants (RAs) conducting fieldwork for this project are First Nations people. While Collard, our senior research assistant, is also a project leader, three additional RAs were also hired.
This team includes three Nyungar fieldworkers: Hope Narrier, Katrina Cox
and Lily Hayward. While Narrier was introduced to the
leadership team through one of Rodríguez Louro’s PhD students,
Cox and Hayward were recommended to the project team
by UWA’s School of Indigenous Studies. This recommendation was
made after Rodríguez Louro contacted the School’s administration
with an offer to accept Expressions of Interest for the positions of
research assistants, which Cox and Hayward were successful
in obtaining.

The maintenance of relationships with Indigenous stakeholders is critical (\S\ref{sec:projects_with_communities}). Rodríguez Louro and Collard are in daily contact via phone messaging, a practice that dates back to 2018 when they began their research together. Rodríguez Louro also maintains regular contact with the Indigenous RAs and quarterly direct contact with the Advisory Committee, although contact on social media is more frequent.

Additionally, outreach on the project to Indigenous communities facilitates greater accountability through public visibility. The project has been reported on by Indigenous media outlets \cite{DudleyNicholson2025, IBN2025}, and Collard has been interviewed about the project on national radio stations \cite{Collard2025}. Collard, Rodríguez Louro and Hutchinson will also jointly present on aspects of the project at the PULiiMA 2025 Indigenous Languages and Technology Conference in Darwin, Northern Territory, a forum which expects over 700 Indigenous attendees.

For convenience, in Appendix~C we re-iterate and summarise aspects of the projects which mitigate the risks and facilitate the opportunities that we enumerated in \S\ref{sec:risks and opportunities}.

\subsection{Limitations}

The technological goals of this project were set by \google{}, and this precluded Indigenous people exercising full control over the project goals. In other words, this project was not driven by or from the community, limiting the opportunities for self-determination. Although innovating on method, and embracing alternate methodologies, the project is nevertheless situated within the dominant paradigm of NLP that seeks to address technical shortcomings through richer datasets, rather than other paradigms which focus on local human agency \cite{bird2024must}. We also acknowledge that First Nations people in Australia are less likely to be in full time research positions at companies or universities in Australia, limiting opportunities to be partner investigators on this research.

We acknowledge the plurality of Indigenous cultures and Countries in Australia, and that Indigenous co-investigators, designers,  managers and transcribers cannot represent the full diversity of those cultures and Countries. Our decision to include multiple regions is important not only because of lexico-grammatical differences, but also because Australian Aboriginal English has a key emblematic function as an encoder of different ways of being and doing.
We believe that the various ancestries represented wherever Australian Aboriginal English is used are as important as understanding Australian Aboriginal English itself.

Our corpus for developing the initial version of the transcription guidelines was limited to publicly shared YouTube videos. We do not claim that it is representative of spoken Australian Aboriginal English in other domains, such as personal messages, nor of written forms of Australian Aboriginal English.   

As researchers based in Australia, our lived experiences with colonial contexts are also based primarily in Australia. Although we hope that this project contains aspects that generalise, we do not claim to know or represent other contexts. 
\section{Discussion: Further Technology Opportunities for Australian Aboriginal English}
\label{sec:opportunities}


We now consider technologies supporting Australian Aboriginal English speakers beyond ASR. Given that languages go by multiple names, the NLP community typically uses language codes as identifiers in multilingual systems, datasets, and data catalogues \cite{kreutzer-etal-2022-quality}. This discrete mapping of languages to codes is central to NLP's conceptual model of what languages are \cite{smart2024socially, bird2024must}. Australian Aboriginal English currently lacks a standardised identifier, and this would be useful for future work on this variety. However, given the power of classifying \cite{bowker2000sorting}---and the colonial associations of the origins of the ISO~639-3 identifiers \cite{dobrin2009practical}, which also underpin codes such as the BCP~47---we echo calls for involving community members and linguists, and avoiding offensive exonyms, when designing codes for Australian Aboriginal English \cite{morey2013language}.

Given the primary orality of Australian Aboriginal English, and its distinct grammatical and discourse-pragmatic features relevant to information seeking (\S\ref{sec:aae}), there are open questions around whether spoken language assistants could better support speakers. Robust support for tag-questions---rather than just questions with subject-verb inversion---might provide more natural support for information requests. While speakers of Australian Aboriginal English benefit from hearing others talk in the same way \cite[p.\ 42]{angelo2021countering}, the question of whether digital assistants should try to `sound Indigenous,' must be approached with care, to avoid potentially reproducing interactions of racialised subservience to non-Indigenous consumers \cite{owens2023}.

Before developing writing tools tailored for Australian Aboriginal English, more understanding is required of how writing practices diverge from standardised Australian English. Social media and other domains with informal registers are more likely to have written Australian Aboriginal English, and it may be that modeling orthographic variation is a useful approach for text technologies \cite[cf.\ ][]{lin2024modeling}.

Evidence suggests that designing health campaigns in Australian Aboriginal English has community benefits \cite{collard2025spark}. This suggests potential use cases for machine translation, particularly automated translation and rewriting of health and other social services materials in Australian Aboriginal English.

Finally, given the prevalence of switching between Australian Aboriginal English and Standardised Australian English, technologies which support the former would benefit from also supporting the latter. Technologies which aim to support English in Australia should explicitly document whether they support Australian Aboriginal English \cite{bender2019benderrule}.

\section{Conclusions}

This paper has explored three questions.  First, can speech technology projects support speakers of Australian Aboriginal English by affording specific opportunities? Second, what are the system deployment and procedural risks of such endeavours? Third, how can we mitigate risks, and facilitate opportunities, by integrating participatory practices which appropriately respect the context and are culturally appropriate for the community?
We explored these questions through a detailed study of the design considerations of a real-world project that aims to improve ASR for Australian Aboriginal English.
The case study of ASR also brings greater attention to a more general trend: the lack of technological support for Indigenous contact language varieties, and the ways that this perpetuates neo-colonialism through a continued failure to recognise and support Indigenous socio-cultural identities and ways of being. Contact language varieties are critical to contemporary Indigenous sociolinguistic identities, and to civil and economic participation, including in places such as Australia where settler-colonialism has led to the extinction of most ancestral languages. We discussed a number of ways in which speech technologies can support speakers of Australian Aboriginal English, as well as the risks of harms caused by system errors.
This paper contributes to our understanding of ethical and responsible development of technologies for minoritised language communities, as well as to our understanding of practices for building culturally sensitive  technologies while incorporating participatory methods. Although the project focuses on a specific technology, community and context---ASR for speakers of an indigenised variety of English in what we now know as Australia---we believe that some of the considerations and methodologies may be applicable to other technology projects.
\section{Ethical considerations}
\label{sec:considerations}

The collection of Australian Aboriginal English for use by \google{} must be approached with extreme care. From the start, \celeste{} and \glenys{} clarified with their partners that they would not use unstructured yarning sessions like they had done for RA/4/20/4977 ‘Aboriginal English in the global city: Minorities and language change’ (see, \eg{}, \cite{rodriguez2021working}). With no pre-determined questions and clear Indigenous leadership through the research process, that project collected language as used in everyday settings, including sensitive stories about dispossession, trauma and suicide. These materials are firmly closed access with no risk of the audio files and associated transcriptions entering the public domain. However, unstructured yarning sessions tend to give rise to highly culturally sensitive material, which should be protected from use by third parties, especially large technology companies like \google{}. For this reason, in consultation with \ben{} and the Advisory Committee, \celeste{} and \glenys{} recommended for this project the design of conversational prompts and story boards (see Appendix~A) which seek to elicit interaction on a variety of general topics while steering clear of culturally sensitive material. 


Only publicly shared YouTube videos were included in the  corpus for developing transcription guidelines, and the data was not copied, modified or re-published, however the video creators were not contacted in order to understand their sentiments towards this research.

\section{Positionality}

Collard, Hutchinson, and Cooper are born in Australia, and \celeste{} is an Australian born and raised in Argentina. Together, we have ancestral connections to Australian Indigenous Countries, Europe and South America, and we have trained and/or worked in linguistics, education, law, natural language processing, software engineering, and cybernetics.  
\glenys{} is an L1 speaker of Australian Aboriginal English, and \glenys{} and \celeste{}  have been active in advocating for greater appreciation of Australian Aboriginal English in civil society.

\begin{acks}
    
\uwa{}, where the second and third authors are based, is situated on unceded Nyungar Country; Google's Sydney office, where the first author is based, is situated on unceded Gadigal Country; and The Australian National University's Acton campus, where the last author is based, is situated on unceded Ngunnawal and Ngambri Country. All authors acknowledge that the Nyungar, Gadigal, Ngunnawal and Ngambri people respectively remain the spiritual and cultural custodians of their Country, on which they continue to practise their values, languages, beliefs, and knowledge.

We would like to thank the Advisory Committee members Prof.\ Clint Bracknell and Mx Sharon Davis for their guidance; research assistants Hope Narrier, Katrina Cox and Lily Hayward for their data collection with members of their community; Emily Gilchrist for her help with project administration and data processing; Nefeli Perdikouli, Mitch Browne, Grace Shepherd, Madeleine Clews
and Lucia Fraiese for their help identifying Aboriginal English in YouTube videos; and Vera Axelrod, Daan van Esch, Prof.\ Michael Haugh, Robert McLellan and Scott Riddle for their project support and advice. We would also like to thank Steven Bird and the anonymous reviewers of FAccT and other venues for their questions and suggestions.

\end{acks}

\bibliography{main}
\bibliographystyle{ACM-Reference-Format}

\newpage

\clearpage
\onecolumn

\section*{Appendix A: Visual prompts for eliciting speech data collection}
\begin{figure*}[h]
    \centering
\subfigure[Food]{\includegraphics[scale=0.1]{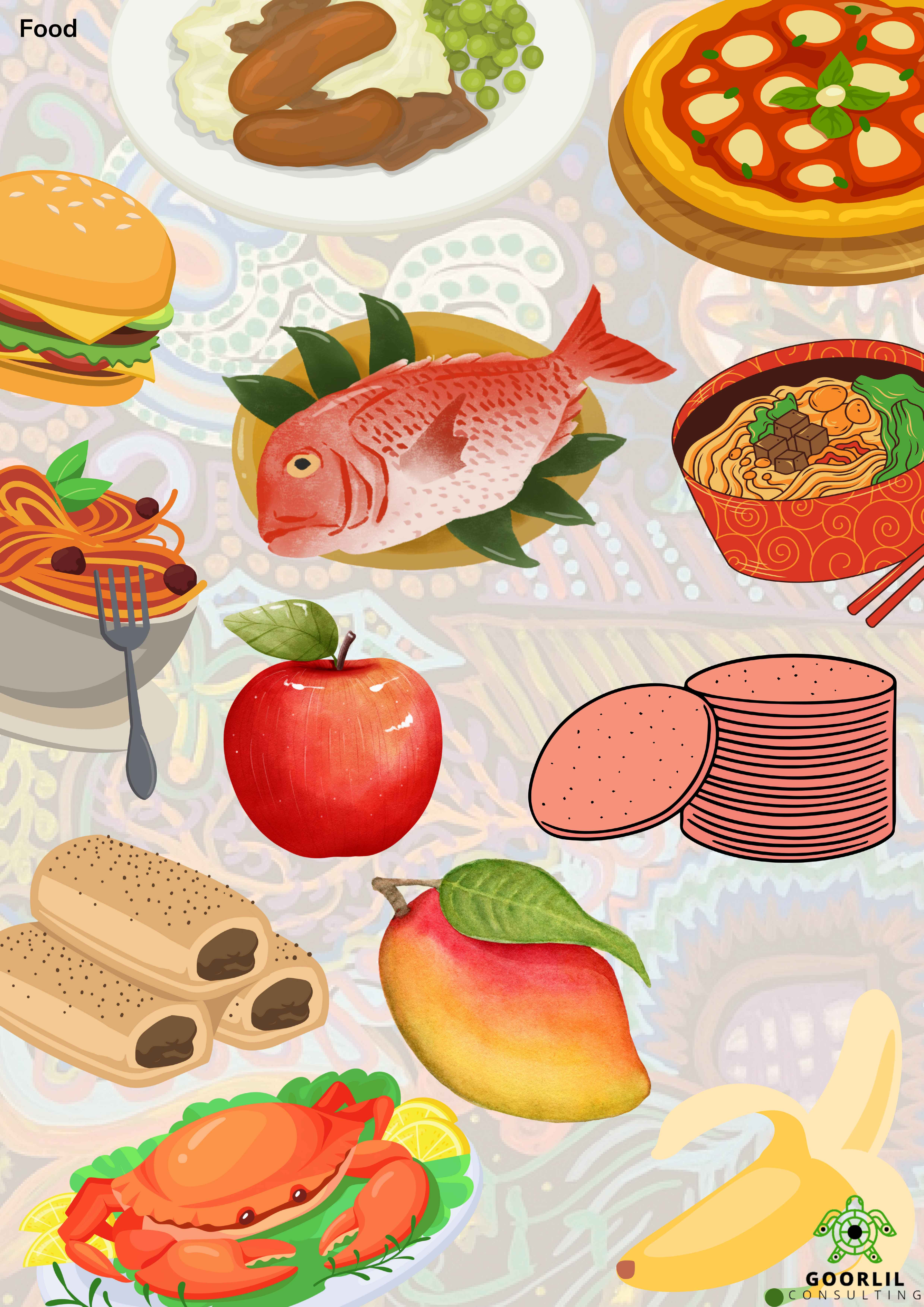}}
\subfigure[Fishing]{\includegraphics[scale=0.1]{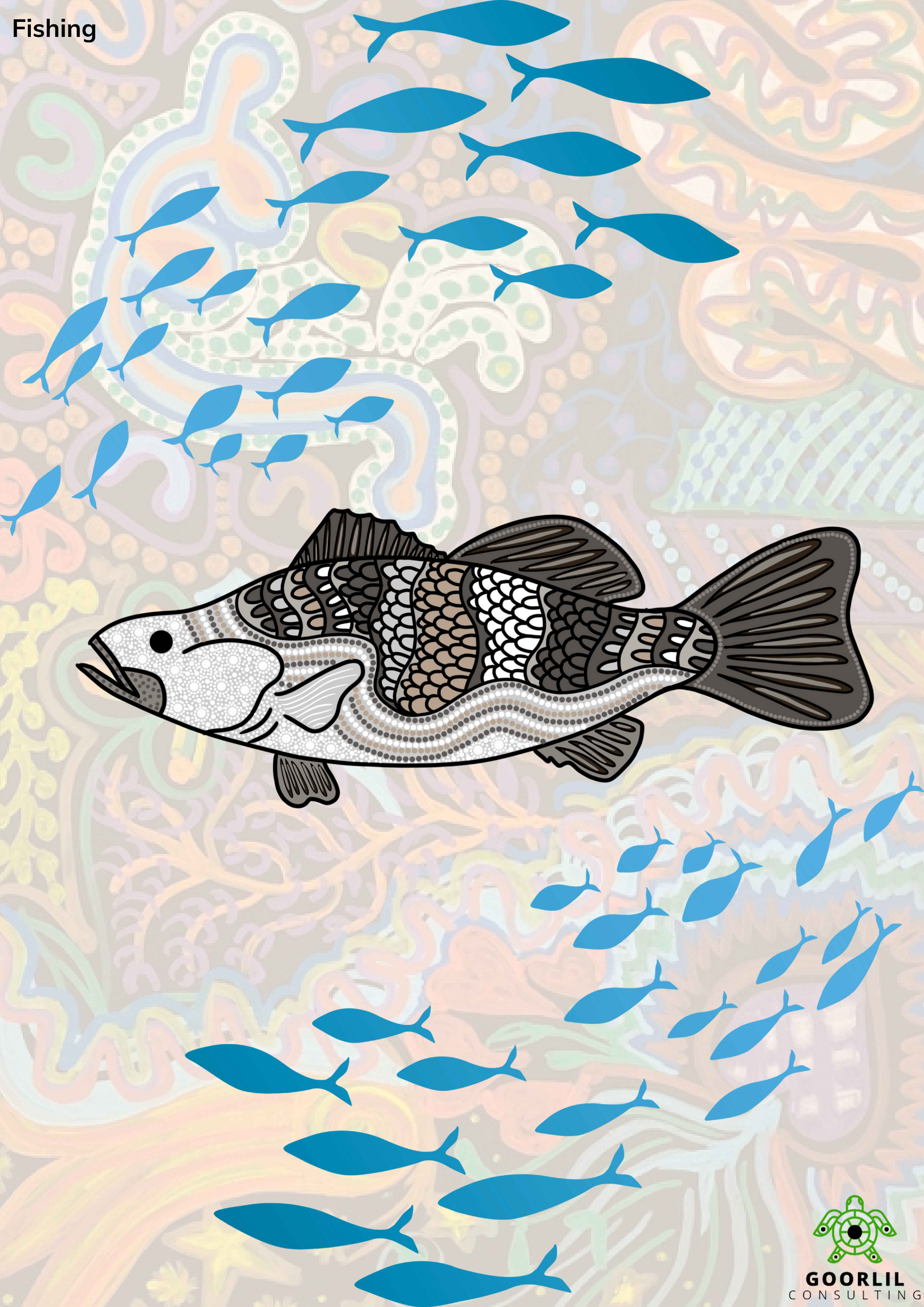}}
\subfigure[Sport]{\includegraphics[scale=0.1]{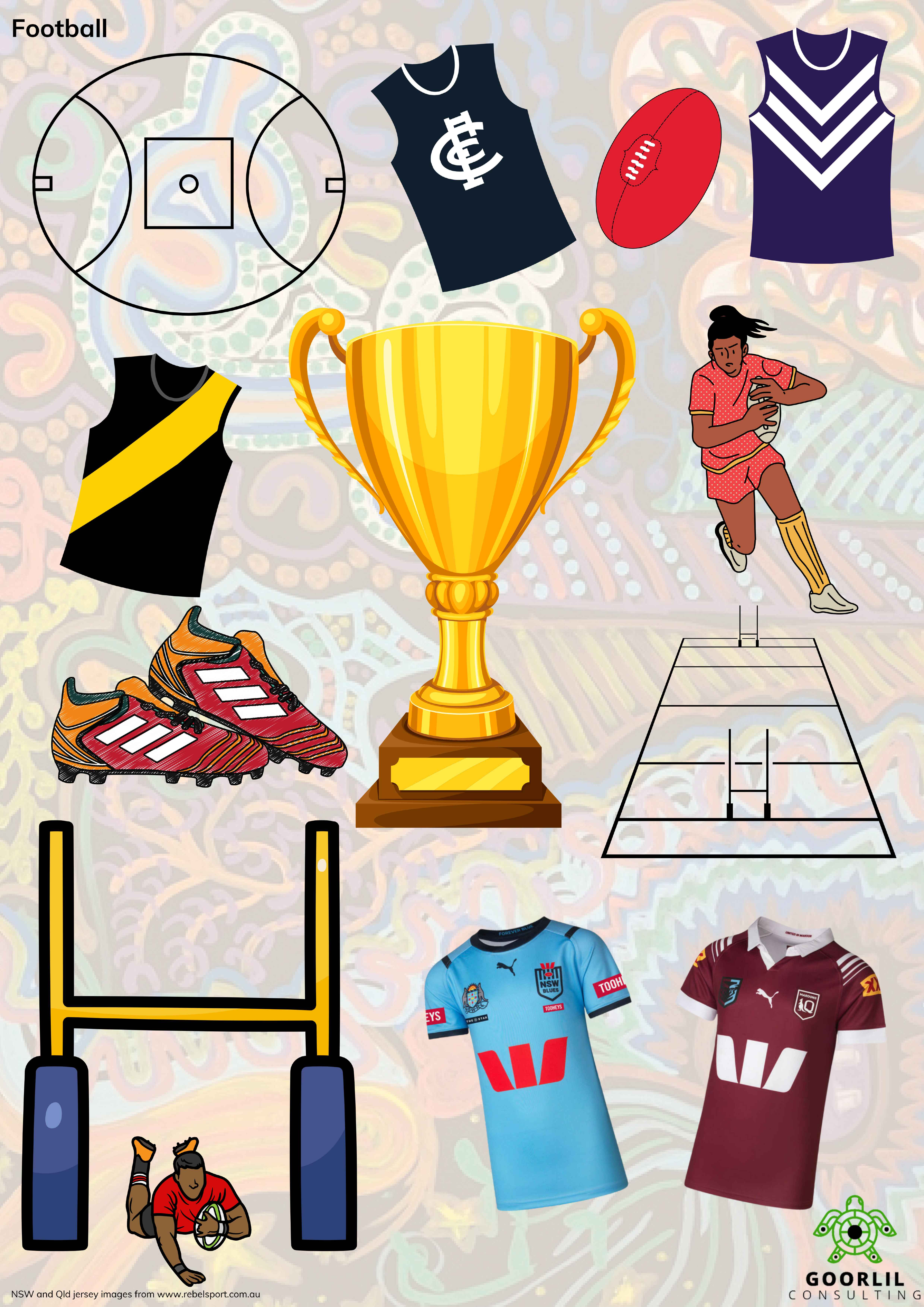}}
\subfigure[Visiting family]{\includegraphics[scale=0.1]{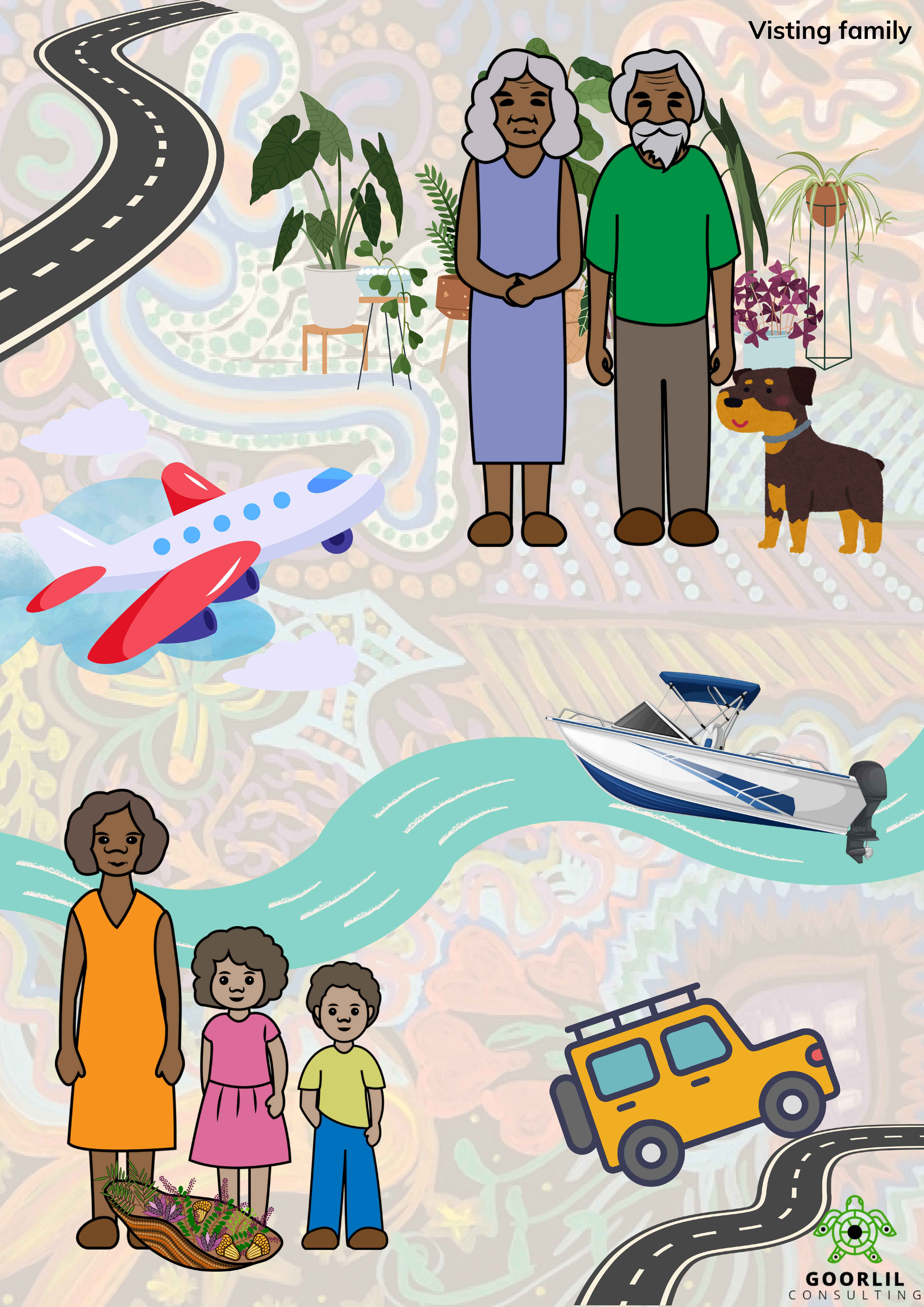}}\\
\subfigure[Hanging out at the shops]{\includegraphics[scale=0.1]{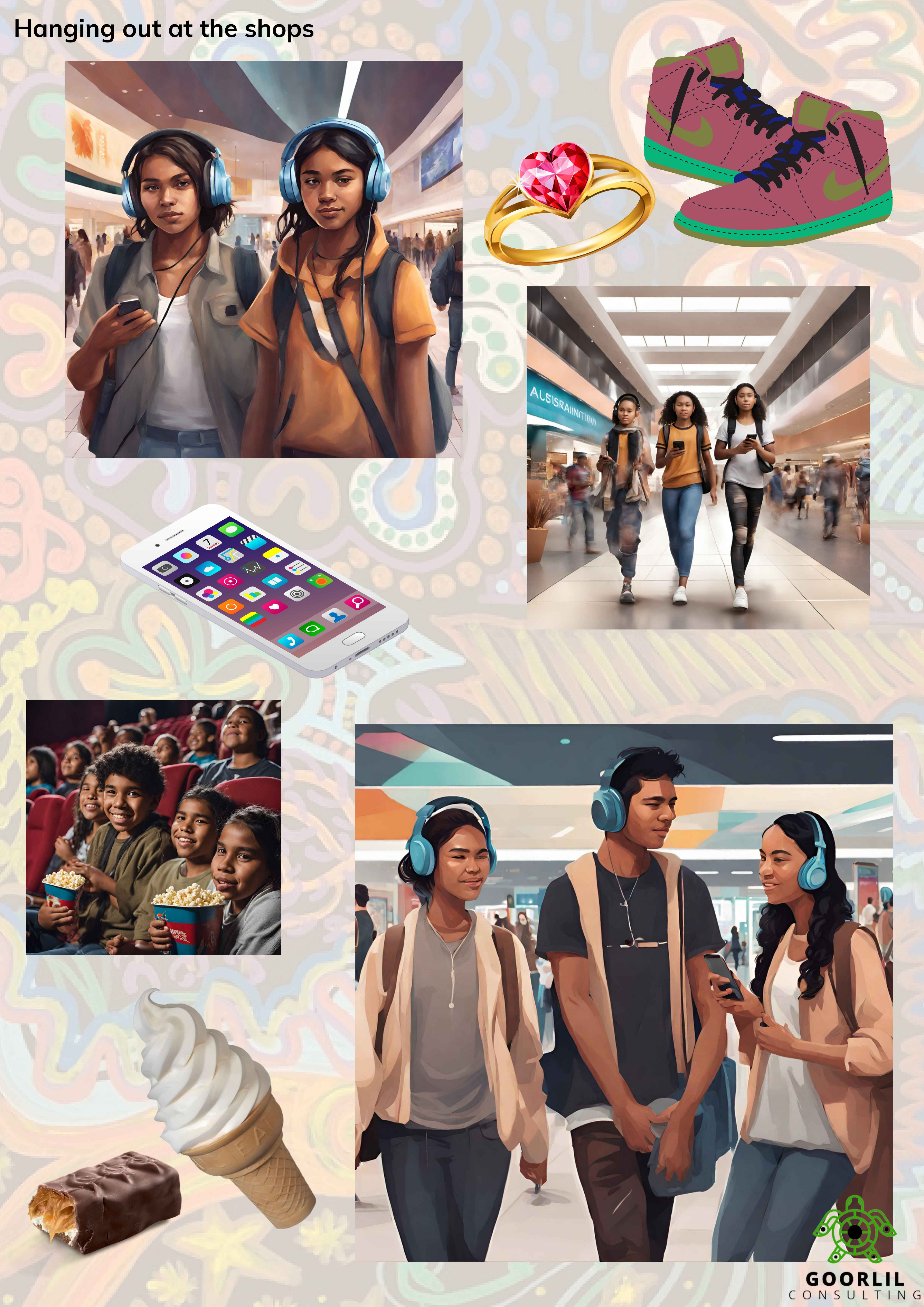}}
\subfigure[Fun times and funny jokes]{\includegraphics[scale=0.1]{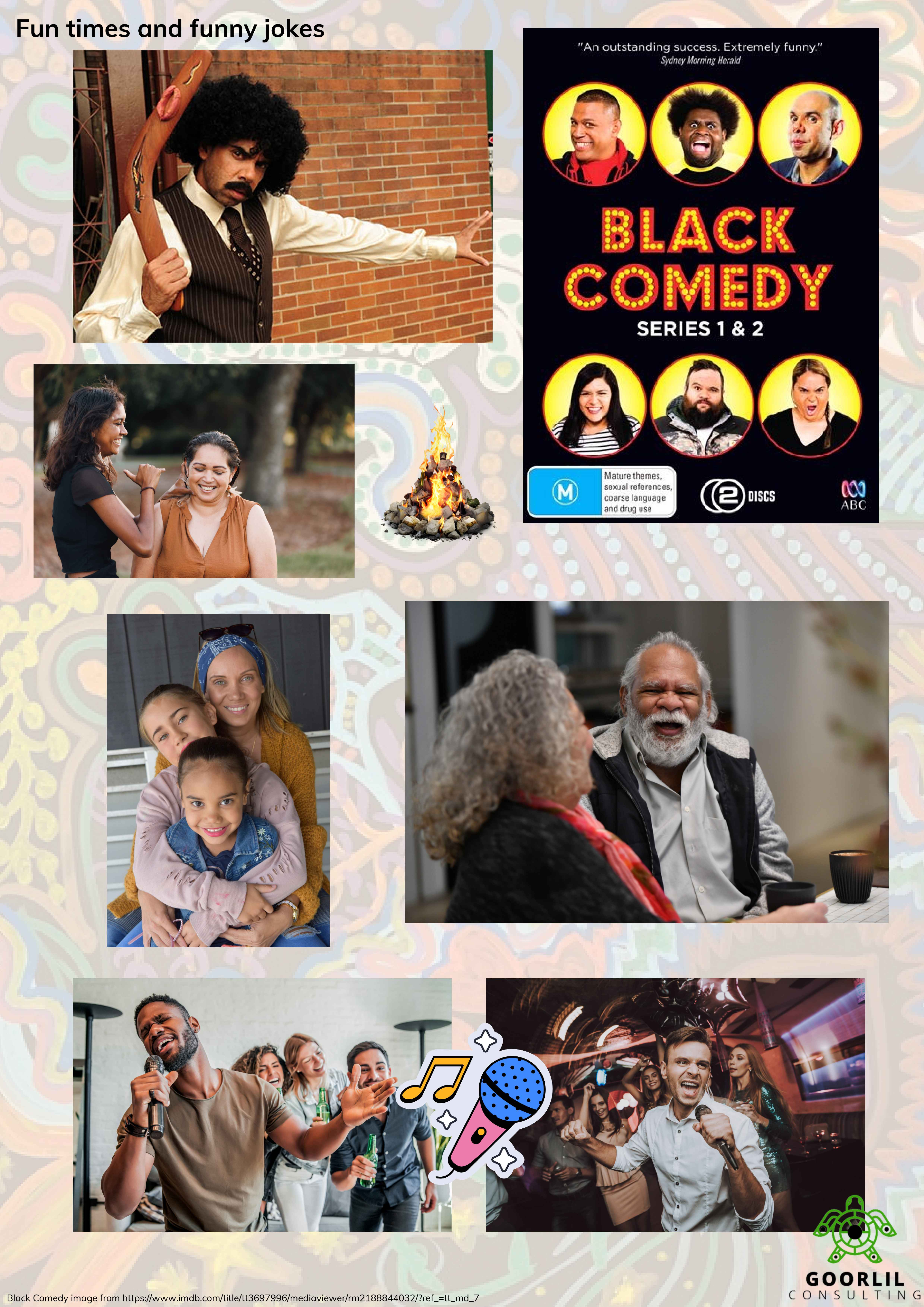}}
\subfigure[Motorcars]{\includegraphics[scale=0.1]{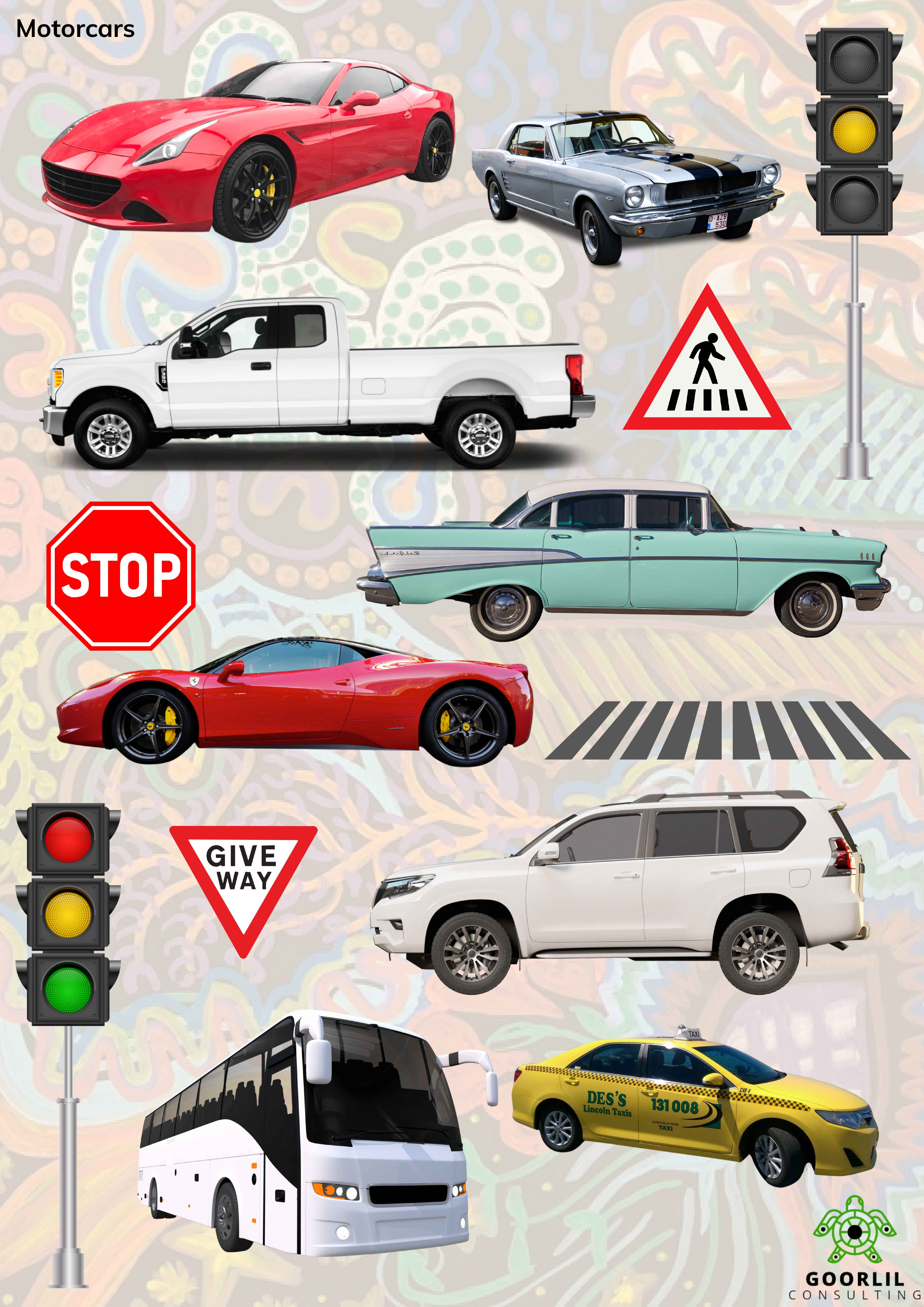}}
    \caption{Visual prompts for speech data elicitation.}
    \label{fig:prompts}
\end{figure*}

The prompts in Figure~\ref{fig:prompts} were designed by Indigenous-owned consultancy Goorlil Consulting.

\clearpage

\twocolumn

\section*{Appendix B: Composition of corpus of Australian Aboriginal English on  YouTube}

For readers who wish to hear examples of Australian Aboriginal English, we recommend searching on YouTube using queries such as ``My art, My culture'', ``Telling Our Stories, Our Stolen Generations'', ``Through our Eyes, Series 3'', or ``Community Spotlight  Artists Profile Cairns Indigenous arts fair''. 

YouTube grants users of the platform `a worldwide, non-exclusive, royalty-free license to access your Content through the Service, and to use that Content, including to reproduce, distribute, prepare derivative works, display, and perform it' (\url{https://www.youtube.com/static?template=terms}). Our use of YouTube videos to create transcription guidelines was consistent with this. These transcription guidelines do not contain any personal identifying information of offensive content.

The YouTube video corpus that we describe in \S\ref{sec:asr} has the composition summarised in Tables 1--3. 

\begin{table}[ht]
    \centering
    \begin{tabular}{p{4cm}rr}
        \toprule
        Domain & \# videos & Total\\
         &  &  duration\\
         &  &  (hours)\\
        \midrule
        Arts and Culture	& 203	& 23.63 \\
Autos and Vehicles	&4	&0.47	\\
Comedy	&29	&2.20	\\
Education		&90	&14.38\\	
Entertainment		&16	&2.60\\	
Film and Animation		&30	&7.82\\	
Gaming	&0	&0	\\
Health	&26	&2.53	\\
How to / Style		&14	&1.79\\	
News and Politics		&42	&6.25	\\
Nonprofits and Activism		&102	&14.31\\	
People and Blogs		&149	&22.29	\\
Pets and Animals		&7	&0.82	\\
Science and Technology		&25	&4.00\\	
Sports		&17	&1.91	\\
Travel and Events		&22	&4.37	\\
        \midrule
        Total & 776 & 111.37 \\
        \bottomrule
    \end{tabular}
    \caption{Domains of videos. }
\end{table}
\begin{table}[ht]
    \centering
    \begin{tabular}{lrr}
        \toprule
        Scripted? & \# videos & Total duration (hours)\\
        \midrule
Yes	&	134	&18.31 \\
        No	&	510	& 77.17\\
Unknown	&	132 &	15.89\\
        \midrule
        Total & 776 & 111.37 \\
        \bottomrule
    \end{tabular}
    \caption{Whether videos were scripted. }
\end{table}

\begin{table}[ht]
    \centering
    \begin{tabular}{p{4cm}rr}
        \toprule
        Percent in & \# videos & Total duration\\
        Australian Aboriginal English &  &  (hours)\\
        \midrule
    {} [50, 60)		& 34	& 8.43 \\
    {} [60, 70)		& 34	& 5.77 \\
    {} {[70, 100]}		& 708	& 97.18 \\
        \midrule
        Total & 776 & 111.37 \\
        \bottomrule
    \end{tabular}
    \caption{Percent of speech in Australian Aboriginal English in videos. Only videos with at least 50\% were considered; 70+\% was preferred.}
\end{table}

\newpage
\clearpage
\onecolumn

\section*{Appendix C: Summaries of mitigations of risks and enabling factors of opportunities}

\begin{table*}[h]
\begin{tabular}{p{0.25\textwidth}p{0.7\textwidth}}
\toprule
Risk & Mitigated by \\
\midrule
1. Deploying inappropriate technology & Building on prior work that identified the appropriateness of ASR for this context; Validation through expert partners with expertise working with this community.\\
\addlinespace
2. Low ecological validity & Multi-region data collection to guarantee that regional variations are represented; In-person data collection techniques led by Indigenous employees, so that speakers are comfortable speaking naturally.\\
\addlinespace
3. Psychological harms from ASR errors & Primary goal of the project: reducing ASR errors for Australian Aboriginal English. We also call for more HCI research both on potential harms experienced by Indigenous communities, as well as on potential mitigations in the application layer.\\
\addlinespace
4. Torquing identities & Community involvement in transcription guidelines; We call for more work on the transcription preferences of speakers of non-standardised languages.\\
\addlinespace
5. Harming relationships & Partnering with experts with existing relationships with community; Establishing an Indigenous Advisory Committee.\\
\addlinespace
6. Failing to partner with community & Indigenous researcher on the project; Hiring Indigenous staff; Engaging an Indigenous-owned consultancy; Establishing an Indigenous Advisory Committee.\\
\addlinespace
7. Culturally unsafe methods in data collection & In-person, Indigenous-led data collection; Using culturally safe prompts developed by an Indigenous-owned consultancy; Verifying collected data for cultural safety; Researchers on the project with experience running culturally safe data collections.\\
\addlinespace
8. Not respecting Indigenous data sovereignty & \google{} not owning the collected speech data; Indigenous Advisory Committee guidance on appropriate Indigenous Data Governance.\\
\addlinespace
9. Not acknowledging Indigenous knowledge and expertise & Recognising contributors as co-authors on research artefacts; Engaging an Indigenous-owned consultancy; Establishing an Indigenous Advisory Committee.\\
\bottomrule
\end{tabular}
\caption{Project risks and mitigations.}
\label{tab:risks}
\end{table*}

\begin{table*}[h]
\begin{tabular}{p{0.35\textwidth}p{0.6\textwidth}}
\toprule
Opportunity & {Enabled by} \\
\midrule
1--4. Deployed system helpful to community and non-community members in diverse application contexts & Explicitly including support for various application contexts in design desiderata.\\
\addlinespace
5. Project members develop greater understanding of Indigenous contexts and histories & Team discussions on power dynamics, barriers, and social injustices; Team discussions about language use during the writing of this paper and other artefacts.\\
\addlinespace
6. Economic opportunities & Equitable payment for Indigenous data providers; Employing Indigenous staff.\\
\addlinespace
7. Building capabilities & Employing Indigenous staff in roles that enable skill development, including designing elicitation materials and transcription guidelines, and managing voice data collection.\\
\addlinespace
8. Self-determination & Pursuing data governance frameworks which give Indigenous people the power to decide how the Indigenous voice recordings might be used in the future.\\
\bottomrule
\end{tabular}
\caption{Project opportunities and enabling factors.}
\label{tab:opportunities}
\end{table*}

\end{document}